# Predictive linguistic cues for fake news: a societal artificial intelligence problem

**Sandhya Aneja[1], Nagender Aneja[2], Ponnurangam Kumaraguru[3]**
[1]Faculty of Integrated Technologies, Universiti Brunei Darussalam, Gadong, Brunei Darussalam
[2]School of Digital Science, Universiti Brunei Darussalam, Gadong, Brunei Darussalam
[3]Department of Computer Science and Engineering, IIIT Hyderabad, New Delhi, India

**ABSTRACT**

Media news are making a large part of public opinion and, therefore, must not be fake. News on web sites, blogs, and social media must be analyzed before being published. In this paper, we present linguistic characteristics of media news items to differentiate between fake news and real news using machine learning algorithms. Neural fake news generation, headlines created by machines, semantic incongruities in text and image captions generated by machine are other types of fake news problems. These problems use neural networks which mainly control distributional features rather than evidence. We propose applying correlation between features set and class, and correlation among the features to compute correlation attribute evaluation metric and covariance metric to compute variance of attributes over the news items. Features unique, negative, positive, and cardinal numbers with high values on the metrics are observed to provide a high area under the curve (AUC) and F1-score.

*Corresponding Author:*

Nagender Aneja
School of Digital Science, Universiti Brunei Darussalam,
Academy of Brunei Studies, BE1410, Brunei Darussalam
Email: nagender.aneja@ubd.edu.bn

## 1. INTRODUCTION

In the media, fake news refers to news that has been fabricated and is presented to readers as being accurate. People in advanced economies are likely to see to more fake content (70%) than real content. Fake news can now be created by humans or by artificial intelligence (AI) [1]. There are numerous fact-checking tools available, including NewsGuard and Hoaxy. Fact-checking websites, such as PolitiFact, GossipCop, and BuzzFeed [2], are still working on improving their ability to identify false information. However, while the quality of news content on social media is lower than that of traditional media, around 50% of Americans in 2021 get news from social media [3]. The revenue from traditional news media is also shrinking, and the online publishers are trying to earn advertising revenue by having more clicks on their content. The distrust of facts proffered by the established media is also rising. Because of the rapid dissemination, easy access, and low-cost dissemination of news on social media, the number of fake news stories is increasing all the time [4].

The goal of the linguistic analysis is to look for language leakage, also called predictive linguistic cues to detect fake news. Recent work on automatic detection captures the predictive cues or writing style using linguistic features, e.g., lexical, syntax, semantic features of the fake content [5], [6]. The news writing style captures the frequency of words accounted in content at linguistic-level, choice between noun/pronoun, writing cardinal number (CN), adjectives, using verbs at syntax level, and psycho-linguistic attributes at the semantic level. Writers of fake news prefer to use their language strategically to influence human psychology.

Rashkin *et al.* [7] presented that language stylistic cues can determine the truthfulness of text. The authors compared the language of real news (from English Gigaword corpus) with that of satire (The Onion, The Borowitz Report, Clickhole), hoaxes (American News, DC Gazette), and propaganda (The Natural News, Activist Report). The authors observed lexicon markers e.g., swear, 2nd person pronoun, modal adverb, action adverb, 1st person pronoun singular, manner adverb, gender, see, negation, strong subjective, hedge, superlatives, weak subjective were more prominent in fake news, while number, hear, money, assertive, and comparatives were more prominent in the truthful news. The fake news detection algorithm is further shown to depend on the stance classification of a news [8].

Allcott and Gentzkow [9] studied news articles of the 2016 US elections. They collected 156 news articles from which 41 were recorded as anti-Trump and 115 as anti-Clinton. Anti-Clinton articles were found 30.3 million times shared on Facebook. The sentiment analysis in [10], [11] including the positive and negative sentiment of input text for news classification seems promising. A study by Horne and Adali [10] on the headline concerning text-body of news for stance classification concluded that headline in fake news repeats the main content.

Efforts are being made to automate the process of fake news detection [12]–[16]. One such technique is Generating aRticles by Only Viewing mEtadata Records (GROVER) [17] which generates fake news and then uses nucleus sampling at each time step to sample from the most probable words whose cumulative probability comprises the top-p% of the entire vocabulary, to create fake news. It gives around 92% accuracy. However, when it is applied to human written fake news it gives 73% accuracy. Thus, it is required to create classifiers that are trained on language written by humans. Sentences created by generative models are distinguishable from human generated text due to the property of low variance and small vocabulary. This property is used by descriptors to the validity of the text [18]. The success of machine learning models depends on feature engineering since all features of a dataset might not be useful in building a machine learning (ML) model for prediction [19]–[23]. Accurate selection of effective features is a crucial step for applying ML algorithms. The automated approach given by Maronikolakis *et al.* [24] applies many recurrent neural networks (RNN) models to detect headlines created by humans or machine generated news. The paper analyses human and machine generated headlines. It was found that humans were only able to identify the fake headlines in 45% of the cases, whereas, the most accurate automatic approach of transfer learning in the paper achieved an accuracy of 94%.

Tan et al. [25] presented an approach to detect the semantic incongruities that are present in text and image captions generated by automated machines. The approach determines the authenticity score by using the co-occurrences of named entities in the text and captions. The word embeddings of captions and image are projected into a common visual semantic space which has a property to be built on fine-grained interactions between words in the caption and objects in the image. A semantic similarity score is computed for every possible pair of projected word and object features. The final authenticity score of an article is determined across those of its images and captions. The approach is compared with GROVER [17] model and outperformed the same. Various deep learning-based techniques are being studied to improve the correlation [26] between features through an attention mechanism. The techniques extend the feature space including multimodal features from audio, video or textual representations into the news content and apply the attention mechanism to mine the complex correlations.

In this paper, fake news detection emphasizes the technique to deeply mine the news content while using the linguistic analysis and language feature set using ML algorithms. We propose applying correlation between features set and class to compute correlation attribute evaluation metric and covariance metric to compute variance over the news items. Proposed feature set can differentiate between fake and real news with high accuracy (nearly $97 \pm 2$ % area under curve (AUC) score) using the AdaBoost model. Main contributions of the paper are:

a) A study of feature set comprising unique words, negative words, neutral words, positive words, compound score, noun, adjective, adverb, preposition, CN for fake news classification.
b) We found a feature set that performed better in comparison to a set of all the features considered in the study using the Corr metric.
c) Results show that the performance of classifiers depend on the news content i.e., linguistic characteristics of the news.
d) Proposed methodology works for balanced, imbalanced, and small datasets.

## 2. METHOD

We used four datasets from Kaggle, BuzzFeed, PolitiFact, and FakeNews Challenge as shown in Table 1. Kaggle-Guardian Dataset comprises fake news from Kaggle and real news from guardian. The Kaggle data set contains text and metadata scraped from 244 different websites tagged as bullshit (BS) by the BS detector chrome extension. We considered only English language news available in the Kaggle dataset. The total number of english language news that was found is 11439. To compare the linguistic features of fake news and real news, we downloaded 9,724 news items from the guardian using guardian application programming interface (API). The news items that we downloaded from the guardian were searched with keywords based on terms in the Kaggle dataset.

BuzzFeed news dataset [2] is collected from fact-checking platform BuzzFeed.com containing news content body text, headline, and uniform resource locator (URL) of the news posted on Twitter by the users. There are 91 real news and 91 fake news propagated through 634,750 social links by 15,257 users. PolitiFact news dataset [2] is collected from fact-checking platform PolitiFact.com similar to Buzzfeed. There are 120 real news and 120 fake news propagated through 574,744 social links by 37,259 users. Fake News Challenge Dataset includes news body text, headline, URL of the news posted with its stance correlated by the user. The dataset has news categorized into four classes agree, disagree, unrelated, and discuss. We changed four classes to two classes by taking agree class as real news while news with stances - disagree, unrelated, and discuss as fake news. There are 49,970 total news with 46,293 fake news and 3,678 real news, this is an imbalanced dataset with 1:12 ratio.

Table 1. Count of news item

|  | Kaggle and Guardian | BuzzFeed | PolitiFact | FakeNews |
|---|---|---|---|---|
| Real News | 9,724 | 91 | 120 | 3,678 |
| Fake News | 11,439 | 91 | 120 | 46,293 |

### 2.1. Feature engineering

To create a feature set, the text of the news was tokenized using word tokenize function of Python nltk library. All tokens that were stop-words as per nltk corpus were removed to create clean text. SentimentIntensityAnalyzer and pos tag were used on the stem of the words from clean text to compute sentiment and parts of speech (POS) tag for each word. SnowballStemmer of Python was used to consider stem of the word to ignore different forms of the word. Frequency of POS tag and sentiment categories were computed for each news item for all the datasets to create features set.

#### 2.1.1. Features set

Features set comprises unique words, negative words, neutral words, positive words, compound score, noun, adjective, adverb, preposition, verb in base form (VB), verb past tense (VBD), verb in gerund or present participle (VBG), verb in past participle (VBN), verb in 3rd person singular present (VBZ), and CN. Unique words represent the number of words that are unique in the given text. Unique words were observed to make 60-100% of fake news while for real news found in the range 20-80%. Positive and negative words represent a measure for identifying the sentiments in the text in terms of intensity and polarity towards emotions [27]. For example, in comparison of two sentences "the person is superb" and "the person is good," the sentence "the person is superb" is considered more sensitive in sentiment intensity analyzer.

Valence aware dictionary for sentiment reasoning (VADAR) [28], a sentiment lexicon available in Python, was used for sentiment analysis. VADAR considers acronyms, initialism like laugh out loud (LOL), emoticons like ;), or slang like nah as crucial for sentiment analysis. The VADAR provides a compound score for intensity scale between -10 to +10. We computed a percentage of negative, positive, and neutral words in both real and fake news. We also considered other grammatical and linguistic features like VB which represents verb base form (for example take), VBD represent sverb past tense (for example took), VBG represents verb gerund/present participle (for example taking), VBN represents verb past participle (for example taken), VBZ represents verb 3rd person present (for example takes) and CN represents cardinal number.

#### 2.1.2. Features selection

Selection of features is based on correlation attribute evaluation metric and covariance metric, which are computed using correlation between features set and class, and correlation among the features. Co-variance of features over the news items is defined by (1) and (2). Let $f_1, f_2, ..., f_n$ represents frequency of all linguistic $n$ features for all $m$ news. We computed $\mu_{real}$ mean frequency of feature $f_i$ over $m_1$ real news and $\mu_{fake}$ mean frequency of feature $f_i$ over $m_2$ fake news. We then calculated covariance of each feature for each real and fake news item as shown in (1) and (2) respectively. We then worked on *Corr* metric to filter the appropriate features. Correlation Attribute evaluation metrics are combined to select the features, so the approach is named as *Corr* [28].

Step 1: Correlation value shows how much one variable changes for a slight change in another variable, and covariance is the direction of the linear relationship between variables. In the proposed method, correlation attribute evaluation metric is evaluated between feature and class ($Corr_{fc}$) averaged over $k$ features. The correlation metric is also evaluated ($Corr_{ff}$) with average over $k$ features and the $Corr$ metric is calculated, the features with high relationship values are selected. If these values are higher than the specified threshold assign value, then the feature is effective, and list is computed in descending order. For evaluation of correlation between the features ($Corr$) correlation between the features set and feature class ($Corr_{fc}$) is calculated. If the correlation between features set and its class is strong, it indicates strong correlation between the features set and class. The wrapper technique is applied to filter the features accurately and select effective features for the selected ML algorithms.

In this technique, features are placed in ascending order with respective correlation values. Afterward, a threshold value is assigned, if feature correlation values are higher than a specified threshold assigned value the feature is put forward in the descending order. We observed that features-unique, positive, negative words and CN are having higher correlation with class and among each other rather than noun, adjective words. Here, we combine (1), (2) and (3) and define correlation and covariance attribute evaluation metrics ($CorrCov$ metric) that is presented in (4) and Table 2.

Table 2. Correlation-covariance attribute evaluation metric

| Attribute | $\dfrac{k_{avg}Corr_{fc}}{nrCovar_{fn} + \sqrt{k + k(k-1)avgCorr_{ff}}}$ |
|---|---|
| A1 | = 0.6 |
| A2 | = 0.4 |
| A3 | = 0.3 |

Step 2: We averaged the co-variance of each feature ($nrCovar_{fn}$) over all news items in the dataset presented in (1) and (2), and after further normalization, the feature was put in a list in descending order. Step 3: Next step is to filter each feature by using the AUC metric of specific ML algorithm. However, the algorithm filters each feature one by one using AUC metric and select those features which give high AUC metric values. The ML algorithms Naive Bayes, decision tree, random forest, K-nearest neighbor, AdaBoost, and support vector machine (SVM) are used to evaluate the AUC metric. Step 4: Final step is verification phase to apply Shannon entropy (using (5)) and technique for order of preference by similarity to ideal solution (TOPSIS) [29], [30] to get desired selected effective feature set in Table 3.

$$Covar_{real_{ithfeature}} = (f_i - \mu_{real_i}) * (f_i - \mu_{real_i}) \tag{1}$$

$$Covar_{fake_{ithfeature}} = (f_i - \mu_{fake_i}) * (f_i - \mu_{fake_i}) \tag{2}$$

$$\dfrac{k_{avg}Corr_{fc}}{\sqrt{k+k(k-1)avgCorr_{ff}}} \tag{3}$$

$$\dfrac{k_{avg}Corr_{fc}}{nrCovar_{fn} + \sqrt{k+k(k-1)avgCorr_{ff}}} \tag{4}$$

$$ent = -\ln(n)^{-1} \sum_{i=1}^{n} A_i \ln(A_i) \tag{5}$$

Table 3. Decision matrix

| Attribute | High-A1 | Medium-A2 | Medium-A3 | Low-A1 | Very High-A2 | High-A3 | Low-A1 | Very Low-A2 | Medium-A3 |
|---|---|---|---|---|---|---|---|---|---|
| Writer1 | 0.7 | 0.5 | 0.4 | 0.3 | 0.9 | 0.7 | 0.3 | 0.1 | 0.5 |
| Writer2 | 0.8 | 0.4 | 0.5 | 0.2 | 0.8 | 0.6 | 0.2 | 0.2 | 0.4 |
| Writer3 | 0.6 | 0.4 | 0.5 | 0.1 | 0.8 | 0.7 | 0.1 | 0.3 | 0.5 |
| ent | 0.651 | 0.954 | 0.954 | 0.834 | 0.458 | 0.715 | 0.834 | 0.834 | 0.954 |
| div | 0.349 | 0.046 | 0.046 | 0.166 | 0.542 | 0.285 | 0.166 | 0.166 | 0.046 |
| wgt | 0.134 | 0.026 | 0.025 | 0.093 | 0.306 | 0.566 | 0.093 | 0.093 | 0.026 |

Different writers have their different writing style of a news while using language attributes (adjectives/adverbs) to write a news. Table 3 is the decision matrix (DM) representing possible different values of selected features. In this research, $A_1 = positive$, $A_2 = negative$, $A_3 = unique$, $A_4 = cardinal number$, $A_5 = variance$ features are found to be effective. Let the features sentiments, nouns, adjectives have range $high = 0.8 - 0.6$, $medium = 0.5 - 0.4$, and $low = 0.3 - 0.1$ values in various news items then the $DM = [\rho_{ij}]_{aXb}$ for different writers is shown in Table 3 where $a$ is the number of writers of news items and $b$ is the number of features. Different classifiers may use different informative feature selection criteria and therefore differ in classification with different weight choices presented in Table 4.

Table 4. Representation for one classifier C1

| Classifier | High-A1 | Medium-A2 | Medium-A3 | Low-A1 | Very High-A2 | High-A3 | Low-A1 | Very Low-A2 | Medium-A3 | Wgt Choice |
|---|---|---|---|---|---|---|---|---|---|---|
| C1 | 1 | 0 | 0 | 0 | 0 | 0 | 0 | 0 | 1 | 0.16 |
| C2 | 1 | 1 | 0 | 0 | 0 | 0 | 0 | 0 | 1 | 0.186 |
| C3 | 1 | 1 | 1 | 1 | 0 | 0 | 0 | 1 | 1 | 0.397 |
| wgt | 0.134 | 0.026 | 0.025 | 0.093 | 0.306 | 0.566 | 0.093 | 0.093 | 0.026 | |

In (5) and (6) provide a quantization of the attributes. The quantization of different classifiers may be used further for training over the datasets by maximization or minimization as in Table 4. Shannon entropy (1-divergence) is the measure of uncertainty and TOPSIS is a statistical method in Table 5 to give ranking of design alternatives. TOPSIS is applied to choose the best solution on the basis of Euclidean distance, shortest distance from the ideal solution (PIS) and the farthest from the negative ideal solution (NIS) in Table 5. Each classifier measures the distance $(\Delta_k^*, \Delta_k^+)$ from PIS and NIS; Therefore, the combined separation distance can be given as: $\Delta_k^* = \sqrt{\sum_{i=1}^n \Delta_{ik}^*}$, $\Delta_k^+ = \sqrt{\sum_{i=1}^n \Delta_{ik}^+}$, where $\Delta_k^* = (wgt_{ik} - MAX(wgt_{ik})^2)$ and $\Delta_k^+ = (wgt_{ik} - MIN(wgt_{ik})^2)$. Each classifier measure closeness of each feature to PIS as $\eta_k^* = \frac{\Delta_k^*}{\Delta_k^* + \Delta_k^+}$. Features obtained using Cov-Corr metric are listed in fSet2 in Table 6.

$$wgt = \frac{div_i}{\sum_{k=1}^n div_k} \quad (6)$$

Table 5. Distance from ideal and negative ideal solution

| Expert | PIS | NIS |
|---|---|---|
| Model 1 | 0.397 | 0.16 |
| Model 2 | 0.697 | 0 |
| Model 3 | 0.06 | 0.1 |

Table 6. Feature sets

| Feature set name | List of features |
|---|---|
| fSet1 | unique, negative, neutral, positive, compound, noun, adjective, adverb, preposition, VB, VBD, VBG, VBN, VBZ, CN, negativeVar, positiveVar, cnVar |
| fSet2 | unique, negative, positive, CN, uniqueVar, negativeVar, positiveVar, cnVar |

## 3. RESULTS AND DISCUSSION

We used two sets fSet1 and fSet2 as presented in Table 6. The fSet1 comprises all features considered to study fake and real news items of four datasets, whereas fSet2 comprises of limited features obtained using Cov-Corr metric. We implemented Naive Bayes, decision tree, random forest, k-nearest neighbors, AdaBoost, SVM algorithms to compare the classification results using fSet1 and fSet2. The scores are obtained by randomly splitting the datasets in the ratio of 0.7:0.3 for the training and cross-validation sets. Figure 1 shows comparison of AUC of algorithms when applied on fSet1. Figure 2 shows comparison of F1 score of algorithms when applied on fSet1. Figure 3 shows comparison of AUC of algorithms when applied on fSet2. Figure 4 shows comparison of F1 score of algorithms when applied on fSet2.

The AUC score is computed from precision-recall curve. We observe high AUC scores for FNC-1 dataset which has more fake news than real news (imbalanced). In Figures 1, 2, 3 and 4, we observe less F1-score ($F1 = \frac{2*Precision*Recall}{Precision+Recall}$) in comparison to AUC score in the classifications by all the ML algorithms. We observe that positive words, negative words, unique words, and CN are the prominent features for the fake news detection from a linguistic analysis of text since the correlation and covariance for the features of fSet2 was higher for fake news than real news as resulted in Figures 1 and 2, however, for rest all other features correlation and covariance was similar. The Figure 1 depicts AUC score obtained using the unique words, negative words, positive words, and CN and the variance of the features (fSet2). In fSet1, we used all the features, however, we could achieve comparable performance with a reduced set of the feature set.

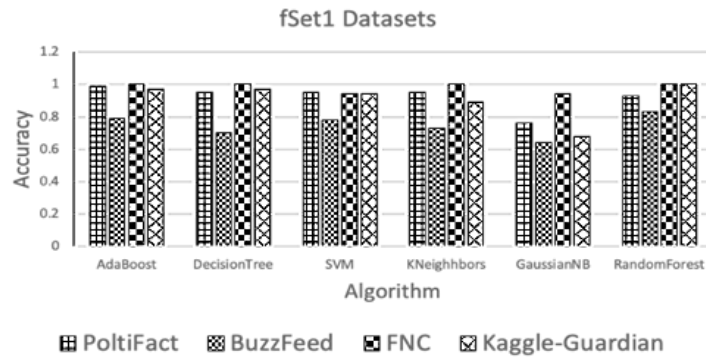

Figure 1. Comparison of AUC score of algorithms on fSet1 with varying datasets

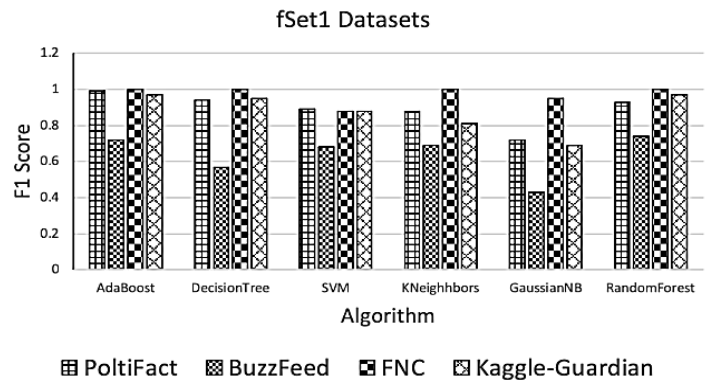

Figure 2. Comparison of F1 score of algorithms on fSet1 with varying datasets

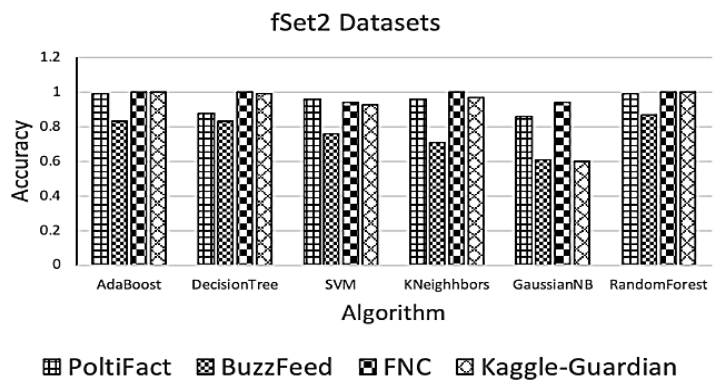

Figure 3. Comparison of AUC score of algorithms on fSet2 with varying datasets

Decision tree algorithm outperformed in comparison to other algorithms except for random forest and AdaBoost. We found that BuzzFeed dataset is the most challenging dataset for all ML algorithms. Decision tree algorithm improved AUC Score from 70% with fSet1 to 83% with fSet2 on BuzzFeed dataset. Gaussian Naive Bayes classifier did not perform well in this particular example of fake/real news identification for all the datasets. Gaussian Naive Bayes classifier uses statistical information of mean and variance of each feature individually over the dataset and then find the joint conditional probability of all features to find the unique range of values for each class. In the FNC -1 dataset which has the highest AUC score with all algorithms, Naive Bayes classifier obtained the AUC score of 94% for fSet1 and fSet2. However, we obtained AUC score of nearly 100% with fSet2 using AdaBoost with base estimator decision tree classifier as shown in Figure 3. One of the reasons may be that naive assumption of gaussian Naive Bayes may not be true since the number of parts of speech depends on each other.

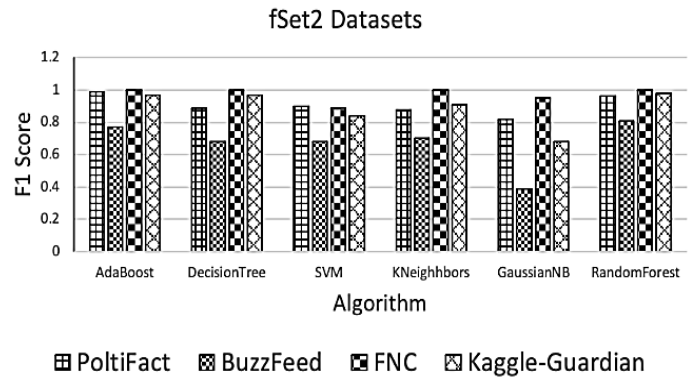

Figure 4. Comparison of F1 score of algorithms on fSet2 with varying datasets

Figure 3 and Figure 4 show that fSet2 outperforms for all the datasets with best performance by Adaboost followed by random forest. There is performance difference of the algorithms on the datasets wherein SVM classifier performed similar for all datasets, including BuzzFeed dataset shown in Figure 3. The SVM classifier is found to be the slowest classifier. We conducted an extensive study to vary all parameters, to find out the best values of hyperparameters for the performance metric. Figures 1, 2, 3 and 4 show F1 and AUC scores for different classifiers and different feature sets fSet1 and fSet2 using the best hyperparameters. We also compared the performance of algorithms when using the best hyperparameters and using the default set of hyperparameters. There was limited performance gain for the algorithms except for random forest classifier and AdaBoost classifier, which improved significantly. The AUC scores of random forest classifier and AdaBoost classifier were improved by 13% and 7% respectively.

Random forest classifier uses subsamples of the feature set to fit into decision tree classifier, and then ensemble obtained trees to predict the class. AdaBoost is a boosting algorithm and is used with week classifiers. In our example with default parameters, it showed the AUC score of 90% with default estimators as decision tree classifier, learning rate 1 and no of estimators as 50. We increased no of estimators to 400 and improved accuracy by 7% leading to 97% AUC score. Now, we present analysis of datasets:

Imbalanced dataset (FNC-1): We observed that fSet2 outperforms in comparison to fSet1 with $97 \pm 2\%$ AUC score. Even though the dataset is skewed, the performance of ML algorithms is up to the mark for both fSet1 and fSet2. Since feature set fSet2 outperformed in comparison to fSet1, therefore we conclude that even though dataset FNC-1 is imbalanced but the frequency of features (e.g. number of unique words, number of positive sentiments words in the news items) was sufficient to perform the accurate classification. We observed that for this dataset fSet1 (other linguistic features in the fake news items) performance is also significant enough due to large numbers of news items (46,293 fake news+3,678 real news) with repeated information for ML algorithms to capture the features from real news and fake news. We observed biased predictions due to imbalance news items in few cases (e.g. the model predicted fake news items with higher accuracy than real news items) and therefore this example presents a scenario of limitation of ML algorithms in avoiding automation of bias [31].

Limited size datasets (PolitiFact and BuzzFeed): The classifiers resulted in low AUC score in comparison to other two datasets. The AUC Score with fSet2 feature set for Buzzfeed is $74 \pm 5\%$. The AUC Score for PolitiFact with fSet2 feature set is $90 \pm 10\%$. Results show that even the datasets are in limited size but the frequency of features in the news is significantly enough therefore the same feature set fSet2 outperformed for the datasets. PolitiFact dataset is better even with fSet1 even though limited in number of real and fake news items.

Balanced dataset (Kaggle-Guardian): Feature set fSet2 in comparison to fSet1 improved the performance for this dataset up to 91 ± 14 % AUC score. Results show that even the dataset is balanced but the frequency of features in the dataset is comparable, therefore, the same feature set fSet2 outperformed for the dataset but with less AUC score than FNC-1. This dataset is difficult for classifiers (less AUC score in comparison to others) though it is balanced.

## 4. CONCLUSION

A study on feature sets over four fake news datasets using ML algorithms conclude that feature set fSet2 is the reduced feature set over the fSet1 since random forest, AdaBoost, k-nearest neighbor, and SVM classifiers obtained high AUC score for fSet2 in comparison to fSet1. The fSet2 is computed using covariance and correlation attribute evaluation metric. The four datasets considered under study were having different proportion of real and fake news items. Thus, the proposed approach has been tested for limited size, imbalanced and balanced datasets. Fake news can be written in regional languages used across the globe to spread the distrust among the local public. Detecting fake news for the regional content is challenging since regional languages have different linguistic features with limited availability of datasets. Future work is proposed over language features for regional languages.

**REFERENCES**

[1] K. D. Stephan and G. Klima, "Artificial intelligence and its natural limits," *AI & SOCIETY*, vol. 36, no. 1, pp. 9–18, May 2021, doi: 10.1007/s00146-020-00995-z.
[2] K. Shu, S. Wang, and H. Liu, "Beyond news contents: the role of social context for fake news detection," in *Proceedings of the Twelfth ACM International Conference on Web Search and Data Mining*, Jan. 2019, pp. 312–320, doi: 10.1145/3289600.3290994.
[3] X. Zhou and R. Zafarani, "Fake news: A survey of research, detection methods, and opportunities," *arXiv preprint*, Dec. 2018, doi: 10.1145/3395046.
[4] F. Monti, F. Frasca, D. Eynard, D. Mannion, and M. M. Bronstein, "Fake news detection on social media using geometric deep learning," *arXiv preprint*, Feb. 2019, doi: 10.48550/arXiv.1902.06673.
[5] X. Zhou, A. Jain, V. V Phoha, and R. Zafarani, "Fake news early detection: a theory-driven model," *arXiv preprint*, Apr. 2019.
[6] N. Aneja and S. Aneja, "Detecting fake news with machine learning," in *International Conference on Deep Learning, Artificial Intelligence and Robotics, (ICDLAIR)*, Springer International Publishing, 2019, pp. 53–64.
[7] H. Rashkin, E. Choi, J. Y. Jang, S. Volkova, and Y. Choi, "Truth of varying shades: analyzing language in fake news and political fact-checking," in *Proceedings of the 2017 Conference on Empirical Methods in Natural Language Processing*, 2017, pp. 2931–2937, doi: 10.18653/v1/d17-1317.
[8] W. Ferreira and A. Vlachos, "Emergent: a novel data-set for stance classification," in *Proceedings of the 2016 Conference of the North American Chapter of the Association for Computational Linguistics: Human Language Technologies*, 2016, pp. 1163–1168.
[9] H. Allcott and M. Gentzkow, "Social media and fake news in the 2016 election," *Journal of Economic Perspectives*, vol. 31, no. 2, pp. 211–236, May 2017, doi: 10.1257/jep.31.2.211.
[10] B. D. Horne and S. Adali, "This just in: fake news packs a lot in title, uses simpler, repetitive content in text body, more similar to satire than real news," *arXiv preprint*, Mar. 2017, [Online]. Available: http://arxiv.org/abs/1703.09398.
[11] R. A. Bagate and R. Suguna, "Sarcasm detection of tweets without #sarcasm: data science approach," *Indonesian Journal of Electrical Engineering and Computer Science*, vol. 23, no. 2, pp. 993–1001, Aug. 2021, doi: 10.11591/ijeecs.v23.i2.pp993-1001.
[12] A. Pardamean and H. F. Pardede, "Tuned bidirectional encoder representations from transformers for fake news detection," *Indonesian Journal of Electrical Engineering and Computer Science*, vol. 22, no. 3, pp. 1667–1671, Jun. 2021, doi: 10.11591/ijeecs.v22.i3.pp1667-1671.
[13] K. M. Fouad, S. F. Sabbeh, and W. Medhat, "Arabic Fake News Detection Using Deep Learning," *Computers, Materials, & Continua*, vol. 71, no. 2, pp. 3647–3665, 2022, doi: 10.32604/cmc.2022.021449.
[14] P. Mookdarsanit and L. Mookdarsanit, "The covid-19 fake news detection in thai social texts," *Bulletin of Electrical Engineering and Informatics*, vol. 10, no. 2, pp. 988–998, Apr. 2021, doi: 10.11591/eei.v10i2.2745.
[15] G. Xiaoning, T. De Zhern, S. W. King, T. Y. Fei, and L. H. Shuan, "News reliability evaluation using latent semantic analysis," *TELKOMNIKA (Telecommunication Computing Electronics and Control)*, vol. 16, no. 4, pp. 1704–1711, Aug. 2018, doi: 10.12928/telkomnika.v16i4.9062.
[16] S. Senhadji and R. A. S. Ahmed, "Fake news detection using naïve Bayes and long short term memory algorithms," *IAES International Journal of Artificial Intelligence (IJ-AI)*, vol. 11, no. 2, pp. 748–754, Jun. 2022, doi: 10.11591/ijai.v11.i2.pp748-754.
[17] R. Zellers *et al.*, "Defending against fake news," *Advances in Neural Information Processing Systems*, vol. 32, 2019.
[18] T. Le, S. Wang, and D. Lee, "Malcom: generating malicious comments to attack neural fake news detection models," in *2020 IEEE International Conference on Data Mining (ICDM)*, 2020, pp. 282–291, doi: 10.1109/icdm50108.2020.00037.
[19] N. Aneja and S. Aneja, "Transfer learning using CNN for handwritten devanagari character recognition," in *2019 1st International Conference on Advances in Information Technology (ICAIT)*, Jul. 2019, pp. 293–296, doi: 10.1109/ICAIT47043.2019.8987286.
[20] S. Aneja, N. Aneja, P. E. Abas, and A. G. Naim, "Transfer learning for cancer diagnosis in histopathological images," *IAES International Journal of Artificial Intelligence (IJ-AI)*, vol. 11, no. 1, pp. 129–136, Mar. 2022, doi: 10.11591/ijai.v11.i1.pp129-136.
[21] S. Aneja, N. Aneja, B. Bhargava, and R. R. Chowdhury, "Device fingerprinting using deep convolutional neural networks," *International Journal of Communication Networks and Distributed Systems*, vol. 28, no. 2, pp. 171–198, 2022, doi: 10.1504/IJCNDS.2022.121197.
[22] S. Aneja, N. Aneja, P. E. Abas, and A. G. Naim, "Defense against adversarial attacks on deep convolutional neural networks through nonlocal denoising," *IAES International Journal of Artificial Intelligence (IJ-AI)*, vol. 11, no. 3, pp. 961–968, 2022, doi: 10.11591/ijai.v11.i3.pp961-968.
[23] S. Aneja, M. A. X. En, and N. Aneja, "Collaborative adversary nodes learning on the logs of IoT devices in an IoT network," in

*2022 14th International Conference on COMmunication Systems & NETworkS (COMSNETS)*, Jan. 2022, pp. 231–235, doi: 10.1109/COMSNETS53615.2022.9668602.
[24] A. Maronikolakis, H. Schutze, and M. Stevenson, "Transformers are better than humans at identifying generated text," *arXiv preprint*, Sep. 2020, [Online]. Available: http://arxiv.org/abs/2009.13375.
[25] R. Tan, B. Plummer, and K. Saenko, "Detecting cross-modal inconsistency to defend against neural fake news," in *Proceedings of the 2020 Conference on Empirical Methods in Natural Language Processing (EMNLP)*, 2020, pp. 2081–2106, doi: 10.18653/v1/2020.emnlp-main.163.
[26] J. Zeng, Y. Zhang, and X. Ma, "Fake news detection for epidemic emergencies via deep correlations between text and images," *Sustainable Cities and Society*, vol. 66, p. 102652, Mar. 2021, doi: 10.1016/j.scs.2020.102652.
[27] T. A. Tran, J. Duangsuwan, and W. Wettayaprasit, "A new approach for extracting and scoring aspect using SentiWordNet," *Indonesian Journal of Electrical Engineering and Computer Science*, vol. 22, no. 3, pp. 1731–1738, Jun. 2021, doi: 10.11591/ijeecs.v22.i3.pp1731-1738.
[28] C. Hutto and E. Gilbert, "Vader: a parsimonious rule-based model for sentiment analysis of social media text," in *Proceedings of the International AAAI Conference on Web and Social Media*, 2014, vol. 8, no. 1.
[29] M. Shafiq, Z. Tian, A. K. Bashir, X. Du, and M. Guizani, "CorrAUC: a malicious Bot-IoT traffic detection method in IoT network using machine learning techniques," *IEEE Internet of Things Journal*, vol. 8, no. 5, pp. 3242–3254, Mar. 2021, doi: 10.1109/jiot.2020.3002255.
[30] T.-C. Wang and H.-D. Lee, "Developing a fuzzy TOPSIS approach based on subjective weights and objective weights," *Expert Systems with Applications*, vol. 36, no. 5, pp. 8980–8985, Jul. 2009, doi: 10.1016/j.eswa.2008.11.035.
[31] D. Varona, Y. Lizama-Mue, and J. L. Suárez, "Machine learning's limitations in avoiding automation of bias," *AI & SOCIETY*, vol. 36, no. 1, pp. 197–203, Jun. 2020, doi: 10.1007/s00146-020-00996-y.

## BIOGRAPHIES OF AUTHORS

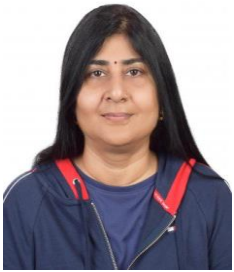

**Sandhya Aneja** is working as Assistant Professor of Information and Communication System Engineering at the Faculty of Integrated Technologies, Universiti Brunei Darussalam. Her primary areas of research interest include wireless networks, high-performance computing, internet of things, artificial intelligence technologies, machine learning, machine translation, deep learning, data science, and data analytics. Further info on her website https://sandhyaaneja.github.io. She can be contacted at email: sandhya.aneja@gmail.com.

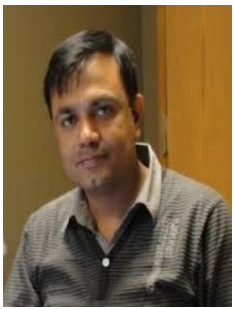

**Nagender Aneja** is working as Assistant Professor at School of Digital Science, Universiti Brunei Darussalam. He did his Ph.D. in Computer Engineering from J.C. Bose University of Science and Technology YMCA, and M.E. Computer Technology and Applications from Delhi College of Engineering. He is currently working in the area of deep learning, computer vision, and natural language processing. He is also founder of ResearchID.co. Further info on his website http://naneja.github.io. He can be contacted at email: naneja@gmail.com.

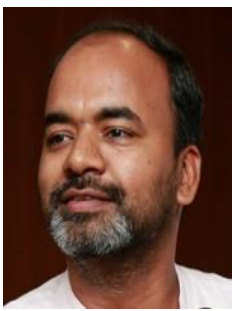

**Ponnurangam Kumaraguru** is a Professor of Computer Science and Dean of Students Affairs at IIIT-Hyderabad. PK is a TEDx and an ACM Distinguished and ACM India Eminent Speaker. PK received his Ph.D. from the School of Computer Science at Carnegie Mellon University (CMU). His Ph.D. thesis work on anti-phishing research at CMU contributed in creating an award-winning startup-Wombat Security Technologies, wombatsecurity.com. Wombat was acquired in March 2018 for USD 225 Million. He can be contacted at email: pk.guru@iiit.ac.in.